%% file: main.tex
\documentclass[conference]{IEEEtran}
\IEEEoverridecommandlockouts

\usepackage{cite}
\usepackage{amsmath,amssymb,amsfonts}
\usepackage{algorithmic}
\usepackage{graphicx}
\usepackage{textcomp}
\usepackage{xcolor}
\usepackage{booktabs}
\usepackage{algorithm}
\usepackage{algorithmic}
\def\BibTeX{{\rm B\kern-.05em{\sc i\kern-.025em b}\kern-.08em
    T\kern-.1667em\lower.7ex\hbox{E}\kern-.125emX}}
\begin{document}

\title{SynSeg: Feature Synergy for Multi-Category Contrastive Learning in End-to-End Open-Vocabulary Semantic Segmentation\\
\thanks{\IEEEauthorrefmark{4}Corresponding author.

This work was supported in part by the National Science and Technology Major Project under Grant No. 2022ZD0114903, by National Natural Science Foundation of China under Grants No. 62302259, 62432008, and 62472248, and by the Talent Fund of Beijing Jiaotong University.}
}

\author{
\IEEEauthorblockN{
Weichen Zhang\IEEEauthorrefmark{1},
Kebin Liu\IEEEauthorrefmark{1}\IEEEauthorrefmark{4},
Fan Dang\IEEEauthorrefmark{2},
Zhui Zhu\IEEEauthorrefmark{1},
Xikai Sun\IEEEauthorrefmark{1},
Yunhao Liu\IEEEauthorrefmark{1}
}
\IEEEauthorblockA{\IEEEauthorrefmark{1}Tsinghua University, 100084, Beijing, China}
\IEEEauthorblockA{\IEEEauthorrefmark{2}Beijing Jiaotong University, 100044, Beijing, China}
\IEEEauthorblockA{\{weic\_zhang23, z-zhu22, sxk23\}@mails.tsinghua.edu, \{kebinliu2021, yunhao\}@tsinghua.edu.cn, dangfan@bjtu.edu.cn}
}
\maketitle

\begin{abstract}
Semantic segmentation in open-vocabulary scenarios presents significant challenges due to the wide range and granularity of semantic categories. Existing weakly-supervised methods often rely on category-specific supervision and ill-suited feature construction methods for contrastive learning, leading to semantic misalignment and poor performance. In this work, we introduce a novel weakly-supervised approach, SynSeg, to address the challenges. SynSeg performs Multi-Category Contrastive Learning (MCCL) as a stronger training signal which robustly injecting intra- and inter-category knowledge during training. We also propose a new feature reconstruction framework named Feature Synergy Structure (FSS). FSS reconstructs discriminative features for contrastive learning through prior fusion and semantic-activation-map enhancement, effectively avoiding the foreground bias introduced by the visual encoder. Furthermore, SynSeg is a lightweight end-to-end solution capable for real-time inference. In general, SynSeg effectively improves the abilities in semantic localization and discrimination under weak supervision in an efficient manner. Extensive experiments on benchmarks demonstrate that our method outperforms state-of-the-art (SOTA) performance, with mIoU score gains ranging from 0.6\% up to 8.9\% across all reported benchmarks. 
\end{abstract}

\begin{IEEEkeywords}
semantic segmentation, open vocabulary, weakly supervised learning, multimodal learning.
\end{IEEEkeywords}

\input{sec/1_intro}
\input{sec/2_relatedwork}
\input{sec/3_method}

\input{sec/4_experiment}
\input{sec/5_conclusion}
\bibliographystyle{IEEEbib}
\bibliography{reference}

\newpage
\input{sec/6_appendix}

\end{document}

%% file: sec/1_intro.tex
\section{Introduction}

Due to the wide range and variability of object categories in the open vocabulary scenarios encountered in real-world tasks, traditional semantic segmentation methods with fixed categories are often insufficient. To overcome these limitations, many Open Vocabulary Semantic Segmentation (OVSS) approaches have been developed recently~\cite{groupvit,pacl,tcl,proxy,cocu,code,ov-seg}. OVSS aims to segment any object category, including those not explicitly defined during training, enabling more flexible and scalable scene understanding. Training such OVSS systems typically requires large amounts of pre-annotated data at pixel level, while semantic annotation in open vocabulary scenarios is both costly and challenging~\cite{cat-seg,openseg}. 
In contrast, weakly-supervised learning methods that incorporate semantic text cues into the semantic segmentation tasks via visual-text alignment techniques offer a compelling alternative.

\begin{figure*}[htbp]
    \centering
    \begin{minipage}[t]{0.2\textwidth}
        \centering
        \includegraphics[height=3.1cm]{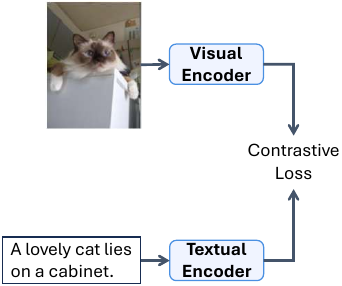}
        \par\vspace{1ex}
        (a) image-text alignment~\cite{groupvit}
    \end{minipage}
    \hfill
    \begin{minipage}[t]{0.31\textwidth}
        \centering
        \includegraphics[height=3.1cm]{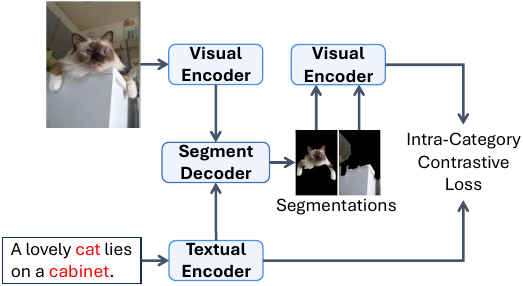}
        \par\vspace{1ex}
        (b) region-text~\cite{tcl} / region-word alignment~\cite{code}
    \end{minipage}
    \hfill
    \begin{minipage}[t]{0.40\textwidth}
        \centering
        \includegraphics[height=3.1cm]{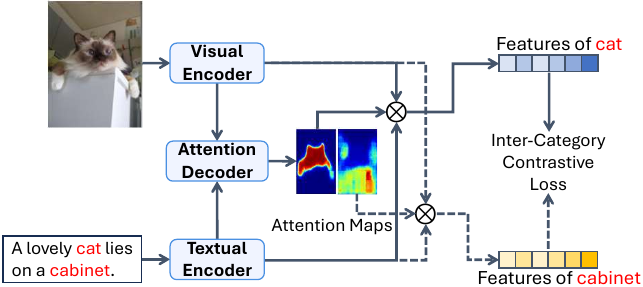}
        \par\vspace{1ex}
        (c) Multi-category separation and alignment (Ours)
    \end{minipage}
    \caption{\textbf{Training paradigms comparison among previous works and ours}. Prior approaches typically adopt either (a) image-text alignment or (b) region-text/region-word alignment, primarily emphasizing \textit{intra-category} contrastive learning. In contrast, our novel paradigm (c) explicitly incorporates \textit{inter-category} contrastive learning for improved discriminative capability. Also, our approach does not need to reconstruct training features from a pre-trained visual encoder.}
    \label{fig:structure}
    \vspace{-1em}
\end{figure*}

Nevertheless, existing weakly-supervised OVSS solutions often fail to achieve satisfactory performance due to the lack of accurate and concrete supervisory signals. GroupViT~\cite{groupvit} is among the earliest approaches in this field and has paved the way for numerous subsequent studies~\cite{s-seg,ren2023viewco}. They utilize a simple image-text alignment architecture shown in Fig.~\ref{fig:structure}(a). Specifically, they operate by grouping local visual features and matching these clusters with text embeddings at test time to generate segmentation masks. However, during training, text embeddings are aligned with a global image representation rather than the detailed region-level features used during inference. This mismatch leads to a notable inconsistency between training and testing, which may impede the model's ability to fully capture fine-grained semantic details.

TCL~\cite{tcl} improved on the above by introducing region-level visual-text alignment training objective, which is shown in Fig.~\ref{fig:structure}(b). Following TCL, some works leverage the directional cues provided by textual descriptions to jointly guide the regions' segmentation in both training and testing process~\cite{code,pacl}. 
These schemes work well in simple scenes with sparse targets. 
They, however, encounter a substantial limitation when applied to scenarios with dense targets which are common in real-word open vocabulary settings~\cite{stuff,coco}. 
This limitation comes from the fact that the training objectives in these solutions are usually designed in a category-specific way, focusing on only intra-category alignment. After all, in real-world scenes where multiple categories appear in close spatial or even visually overlap with each other, intra-category alignment alone is insufficient to achieve strong semantic discrimination. Without a coordinated interplay of both intra- and inter-category alignment and separation mechanisms, the model struggles to disambiguate between neighboring or overlapping regions. Therefore, this leads to the first challenge: \textit{existing OVSS methods lack explicit modeling of inter-category correlations during training, which limits their ability to distinguish semantically different targets in one image.}

Moreover, as shown in Fig.~\ref{fig:structure}(a) and (b), previous works typically rely on frozen, pre-trained vision encoders such as CLIP~\cite{clip} to extract features for contrastive learning \textit{after the decoder}. However, these features often exhibit limited discriminability in background regions, as the encoder is inherently biased toward salient foreground objects~\cite{denseclip}. Therefore, segmentation regions contaminated by background noise may still yield highly similar representations to clean regions. This hinders the rapid decline of the contrastive loss and reduces the overall learning efficiency. As a result, this leads to the second challenge: \textit{existing methods lack the ability to reconstruct representations that are well-suited for contrastive learning in open-vocabulary semantic segmentation.}

To address the two challenges above, our work introduces an innovative approach, SynSeg, for weakly-supervised OVSS. Specifically, we propose a \textbf{Multi-Category Contrastive Learning} strategy, and a new feature reconstruction framework named \textbf{Feature Synergy Structure}.

\textbf{Multi-Category Contrastive Learning (MCCL)} provides a stronger supervisory signal that introduces inter-category alignment and separation across multiple semantic categories, which is shown in Fig.~\ref{fig:structure}(c). MCCL constructs positive pairs between foreground features and their corresponding text embeddings, as well as negative pairs from foreground and background features belonging to the same class. Additionally, it forms positive pairs from background features of these semantic categories, for their backgrounds are often highly overlapped, and negative pairs from different semantic classes within the same image. Our training strategy enhances the distinction between objects in semantic-dense scenarios with explicit inter-category knowledge injection, resulting in more precise semantic localization and segmentation.

To enable effective MCCL, we propose a feature reconstruction framework named \textbf{Feature Synergy Structure (FSS)}. FSS is designed to generate category-aware representations that are more suitable for contrastive learning in open-vocabulary semantic segmentation. In contrast to prior approaches that perform contrastive learning on features directly produced by vision encoders, FSS reconstructs semantic-aware features by explicitly incorporating textual guidance into the feature generation process. By leveraging semantic activation responses, FSS enhances discriminability among dense and overlapping categories while mitigating the foreground bias introduced by pre-trained vision encoders. As a result, the reconstructed features provide a more informative and robust foundation for multi-category contrastive learning.

The solutions we propose effectively address the limitations of current weakly-supervised methods in OVSS and enhance segmentation performance. Our primary contributions include:

\begin{itemize}
    \item We introduce a novel Multi-Category Contrastive Learning (MCCL) strategy that incorporates both inter-category and intra-category contrastive objectives, which provides a stronger weak-supervision signal for OVSS task.
    \item We propose Feature Synergy Structure (FSS) to reconstruct semantic-aware features for effective contrastive learning instead of reusing pretrained visual encoders. The features are enhanced by semantic-activation maps, which emphasize semantically relevant regions while suppressing less informative areas.
    \item We implement and evaluate our proposed method, SynSeg, across multiple OVSS evaluation datasets, achieving performance that surpasses state-of-the-art (SOTA) benchmarks.
\end{itemize}

%% file: sec/2_relatedwork.tex
\section{Related Work}

\subsection{Multimodal Learning \& Visual-Language Models}
Multimodal learning studies how to jointly represent, align, and reason over information from multiple sources or modalities, such as vision, language, audio, and other heterogeneous signals~\cite{mm_survey,WENJUN}. Vision-language models aim to bridge the gap between visual and textual modalities, enabling a broad range of multi-modal tasks such as image-text retrieval~\cite{align,clip} and image captioning~\cite{blip,vinvl}.

\subsection{Weakly-Supervised OVSS Methods}

\begin{figure*}[t]
\centerline{\includegraphics[scale=1.2]{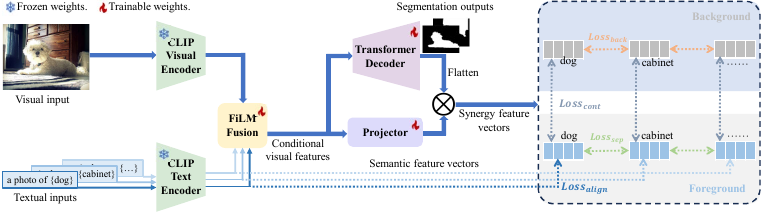}}
\caption{\textbf{The pipeline of SynSeg}. It illustrates the proposed Feature Synergy Structure and Multi-Category Contrastive Learning framework. During training, FiLM~\cite{film} fusion module, transformer decoder and the projector stay trainable, while the CLIP~\cite{clip} encoders stay frozen. The projector is implemented as a convolutional layer to ensure the dimension alignment for feature vectors.}
\label{fig:overview}
\vspace{-1em}
\end{figure*}

In early OVSS works~\cite{groupvit,ren2023viewco}, textual information does not participate in the generation of segmentation masks but only plays a role in the matching of mask proposals with candidate semantic labels. Multiple later works follow this structure and explore improvements in segmentation accuracy and semantic alignment precision~\cite{s-seg,cocu}. The inconsistency between region-level alignment during testing and global-level alignment during training also leads to poor performance in fine-grained segmentation.

TCL~\cite{tcl} performs a different structure firstly to partially incorporate text labels into the mask generation process and apply region-text alignment during training. This paradigm is followed by later works~\cite{pacl,code}. For example, CoDe~\cite{code} improves its work by introducing region-word alignment. In these works, the integration of textual information with the visual feature enables the mask proposal to refer to certain semantic more flexibly, thereby enhancing the granularity of segmentation alignment. However, This training approach mainly focuses on a single object or semantic category and does not exploit the relationships among different semantic objects within the same image. 

Furthermore, these works typically feed segmentation results into the CLIP models~\cite{clip} once again to construct visual features for loss computing. Since the CLIP model is inherently biased toward salient foreground objects, its visual representations tend to be sparse and less discriminative in background regions~\cite{denseclip,maskclip}. In OVSS, the segmentation predictions inadvertently include background noise—pixels that actually belong to other semantic categories—such sparse representations hinder the model’s ability to produce more accurate and precise segmentation regions. 

\subsection{OVSS Based on Large Pretrained Models}
Beyond the above weakly supervised OVSS approaches, a line of recent works follows a training-free paradigm~\cite{proxy,trident,kim2024d}. These methodstypically rely on large-scale pretrained models and refine their intermediate predictions rather than performing fully end-to-end inference, which usually incurs substantial latency and computational overhead. A similar dependency also appears in some weakly supervised methods~\cite{s-seg,dpseg}, which build upon intermediate outputs from powerful pretrained models. For example, ProxyCLIP~\cite{proxy} leverages pretrained segmentation models such as DINO~\cite{dino}, while DPSeg~\cite{dpseg} relies on Stable Diffusion~\cite{sdiffusion} for visual prompting. We acknowledge the value of these explorations that leverage large pretrained models, however, they inevitably sacrifice lightweightness and efficiency. To keep efficiency, we construct our method, SynSeg, in an end-to-end scheme without relying on any large pretrained models. We provide a simple comparison in Table~\ref{tab:speed}, where all results are measured on an RTX 4090 GPU.

\begin{table}[h]
\vspace{-1em}
\centering
\caption{Comparison of model parameters and inference speed on different types of OVSS pipelines.}
\setlength{\tabcolsep}{4pt}
\begin{tabular}{c|c|ccc}
\toprule
Method & End-to-End &Param. & Latency & FPS\\
\midrule
SynSeg (Ours) & Yes&151M &14 ms & 71\\

TCL   & Yes &178M & 13 ms & 76\\ 
\midrule
ProxyCLIP& No& 243M  & 58 ms&17\\ 

DPSeg& No& $\sim$1.2B&- & -\\
\bottomrule
\end{tabular}

\label{tab:speed}
\vspace{-2em}
\end{table}

%% file: sec/3_method.tex
\section{Approach}

\subsection{Feature Synergy Structure}
Our pipeline, named SynSeg, is illustrated in Fig.~\ref{fig:overview}. The pre-trained CLIP~\cite{clip} encoders process the input image and text labels, respectively, to extract single-modal embeddings. These embeddings are then fused using a Feature-wise Linear Modulation (FiLM)~\cite{film} module. Meanwhile, textual embeddings are stored as semantic feature anchors for intra-category alignment later. During inference, the conditional visual features are passed through the decoder, formulating semantic-activation maps. Then, the semantic-activation maps are thresholded to generate the final segmentation outputs. Besides CLIP encoders, SynSeg does not utilize any large pretrained models and follows an end-to-end paradigm, which enables efficient inference.

We introduce a new feature reconstruction framework for training, termed Feature Synergy Structure (FSS), inspired by CCAM~\cite{ccam}. FSS operates on post-decoder features and reconstructs category-aware synergy feature vectors for subsequent contrastive learning. In a single pass, it produces foreground and background feature vectors for each semantic category without reusing the pretrained visual encoder.

The class-specific semantic-activation maps represent the per-pixel response strength to each semantic category and serve as attention maps for feature enhancement. After flattening, they are fused with conditional visual features via matrix multiplication to generate the synergy feature vectors. This process highlights semantically relevant spatial regions while preserving contextual cues. To ensure spatial-semantic consistency, the semantic-activation maps are duplicated and transposed so that each synergy feature vector corresponds to either the foreground or background region of the referred semantic category. The resulting reconstructed features therefore form category-specific foreground and background representations, enabling effective integration of semantic context and fine-grained visual cues.

Throughout the training process, only the FiLM fusion module, the projector, and the transformer decoder parameters are updated, while the CLIP visual and textual encoders remain frozen. This ensures the effective leverage of the rich features from the pre-trained visual language model.

\begin{table*}
\vspace{-1em}
\centering
\caption{\textbf{Zero-shot semantic segmentation comparisons among weakly-supervised OVSS methods.} Bold indicates best performance; underlined values are second-best. Results are in mIoU (\%), which higher is better.}
\begin{tabular}{l|c|c|cccccc|c}
\toprule
    \textbf{Method} & \textbf{Publication} & \textbf{Training Datasets} &
    \textbf{VOC} & \textbf{Context} & \textbf{Object} &
    \textbf{Stuff} & \textbf{City} &\textbf{ADE} &\textbf{Avg.} \\
    \midrule
    GroupViT     & CVPR 2022 & CC3M+CC12M+RedCaps12M   & 50.8 & 23.4 & 27.5 & 15.3 & 11.1 &  9.2 & 22.9 \\
    ViewCo       & ICLR 2023 & CC12M+YFCC14M           & 52.4 & 23.0 & 23.5 &  –   &  –   &  –  & – \\
    CoCu         & NeurIPS 2023 & CC3M+CC12M+YFCC14M  & 51.4 &  23.6  & 22.7 & 15.2 & 22.1 &  12.3   & 24.6 \\
    OVSegmentor  & CVPR 2023 & CC12M                   & 53.8 & 20.4 & 25.1 &  –   &  –   &  –   & –\\
    TCL          & CVPR 2023 & CC3M+CC12M              & 55.0 & 33.9 & 31.6 & 22.4 & 24.0 & 17.1 & 30.7 \\
    CoDe         & CVPR 2024 & CC3M+CC12M              & \underline{57.7} & 30.5 & 32.3 & \underline{23.9} & \underline{28.9} & \underline{17.7} & \underline{31.8} \\
    
    S-Seg        & CVPR 2025 & CC3M+CC12M              & 53.2 & 27.2 & 30.3 &  –   &  –   &  –  & – \\
    MGCCL         & TMC 2025  & CC3M + CC12M                    & 55.1 & \underline{34.9} & \underline{33.0} & 22.8 & 25.9 & 17.3 & 31.5 \\
    \midrule
    SynSeg (Ours)& ICME 2026         & CC12M                   & \textbf{60.5} & \textbf{43.8} & \textbf{36.0} & \textbf{24.5} & \textbf{35.8} &\textbf{18.4} & \textbf{36.5} \\
    \bottomrule
  \end{tabular}
  
  \label{tab:ovss-results}
  \vspace{-1em}
\end{table*}

 \begin{figure*}
   \includegraphics[width=1\textwidth, height=6.5cm]{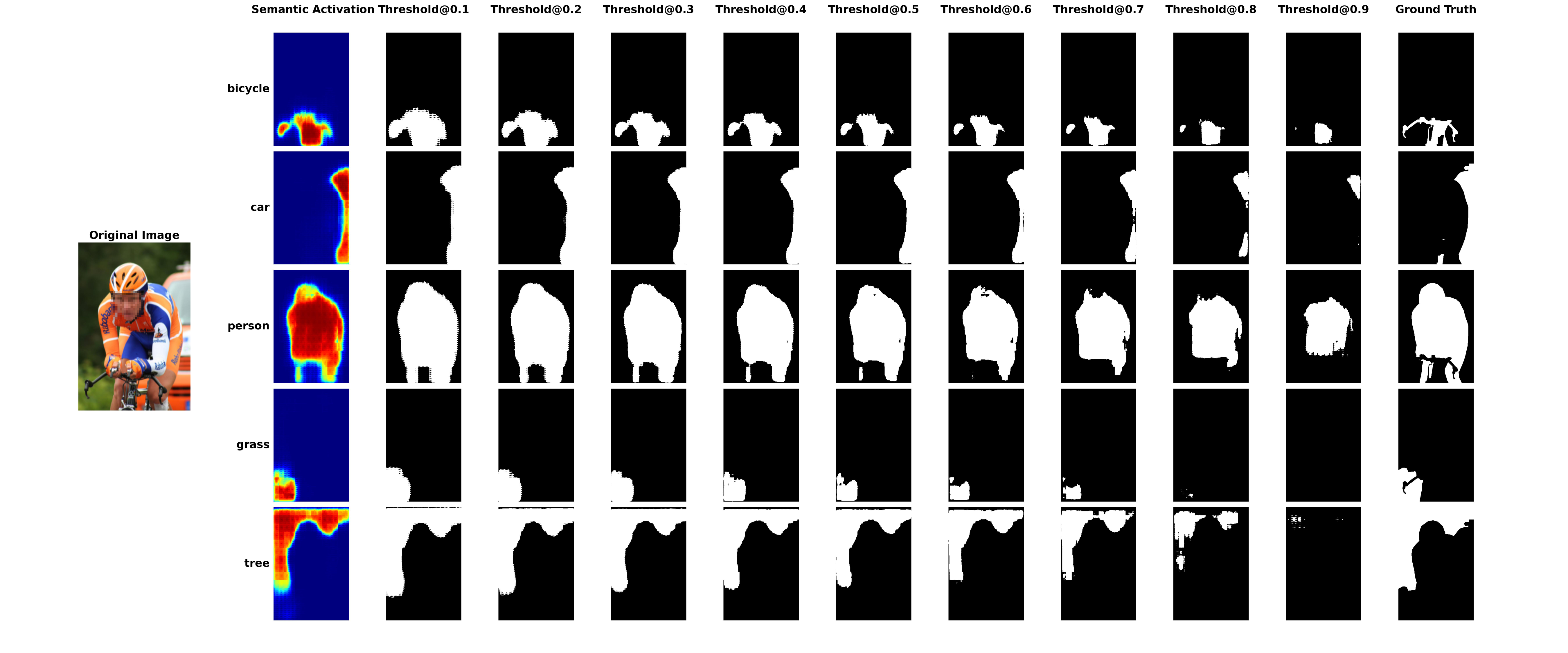}
  \caption{Visual effects of the semantic activation maps and segmentations under different thresholds.}
  \label{fig:sample}
  \vspace{-1em}   
  \end{figure*}
  
\subsection{Multi-Category Contrastive Learning}\label{sec:fs}
We propose Multi-Category Contrastive Learning (MCCL), which models both intra-category correlation and inter-category relations through four contrastive loss terms.

For each category \( c_i \in \mathcal{C}_I \), the model produces a foreground-background pair of synergy feature vectors, denoted by \( f_{c_i} \) and \( \bar{f}_{c_i} \), guided by the semantic-activation map of \( c_i \).The complete sets of foreground and background synergy features for image $I$ can thus be expressed as:
\begin{equation}
\mathcal{F}_I = \{f_{c_i} \mid c_i \in \mathcal{C}_I\}, \quad
\bar{\mathcal{F}_I} = \{\bar{f}_{c_i} \mid c_i \in \mathcal{C}_I\}.
\end{equation}

To incorporate textual semantics, we use the CLIP text encoder to obtain category-level semantic embeddings:
\begin{equation}
\mathcal{T}_{I} = \{t_{c_i} = \mathrm{CLIP}_t(c_i) \mid c_i \in \mathcal{C}_I \},
\end{equation}
where \( t_{c_i} \in \mathbb{R}^d \) serves as the semantic anchor for intra-category alignment.

For all contrastive objectives, we compute cosine similarity between $\ell_2$-normalized features and adopt a bounded form to ensure valid logarithmic inputs. Specifically, for two feature vectors $x,y \in \mathbb{R}^d$, we first compute

\begin{equation}
s(x,y)=\frac{x^\top y}{\|x\|_2\|y\|_2},
\end{equation}
then clamp it into $(-1,1)$ and map it to $(0,1)$:

\begin{equation}
\mathrm{sim}(x,y)=\frac{\mathrm{clamp}(s(x,y),-1+\epsilon,1-\epsilon)+1}{2},
\end{equation}
where $\epsilon=1\times10^{-4}$ is a small constant. Therefore, $\mathrm{sim}(x,y)\in(0,1)$, which ensures numerical stability for all logarithmic terms in the following losses.

To enforce intra-category semantic alignment, the first loss function, \(L_{\mathrm{align}}\), maximizes the similarity between each foreground synergy feature \(f_{c_i}\) and its corresponding semantic embedding \(t_{c_i}\):

\begin{equation}
L_{\mathrm{align}}(\mathcal{F}_I,\mathcal{T}_I) = - \frac{1}{N_I} \sum_{c_i \in \mathcal{C}_I} \log \left( \mathrm{sim}(f_{c_i}, t_{c_i}) \right),
\end{equation}
where \(N_I = |\mathcal{C}_I|\) is the number of semantic categories present in image \( I \).

To enhance the quality of segmentation boundaries, we introduce the second loss function, \( L_{\mathrm{cont}} \), which performs intra-category separation through contrastive learning between foreground and background regions of the same semantic category. For each category \( c_i\), this loss minimizes the cosine similarity between the corresponding foreground and background synergy feature vectors \( f_{c_i} \) and \( \bar{f}_{c_i} \):

\begin{equation}
L_{\mathrm{cont}}(\mathcal{F}_I, \bar{\mathcal{F}_I}) = - \frac{1}{N_I} \sum_{c_i \in \mathcal{C}_I} \log \left( 1 - \mathrm{sim}(f_{c_i}, \bar{f}_{c_i}) \right),
\end{equation}
where \( N_I = |\mathcal{C}_I| \) is the number of categories in image \( I \).

Inter-category alignment loss, \( L_{\mathrm{back}} \), is introduced to constrain foreground over-expansion by aligning complementary background representations across categories within the same image. Since the background of one category naturally contains the regions of other categories, this term acts as a shared ``not-this-class'' regularizer. For a given image \( I \), this loss maximizes the cosine similarity between all background synergy features \( \bar{f}_{c_i}\) associated with each category \( c_i \):

\begin{equation}
L_{\mathrm{back}}(\bar{\mathcal{F}}_I) = 
- \frac{1}{N_I^{\mathrm{pair}}}
\sum_{\substack{c_j, c_k \in \mathcal{C}_I\\ j \neq k}}
\log \left( \mathrm{sim}(\bar{f}_{c_j}, \bar{f}_{c_k}) \right)
\label{eq:backloss}
\end{equation}
where \( \bar{f}_{c_j}, \bar{f}_{c_k} \in \bar{\mathcal{F}_I} \) are the background synergy features for categories \( c_j \) and \( c_k \), and \( N_I^{\mathrm{pair}} = N_I(N_I - 1) \) denotes the number of ordered category pairs with \( j \neq k \) in image \( I \).

To enhance the inter-category separation of individual semantic areas within the same image, we introduce the fourth loss function, \( L_{\mathrm{sep}} \), as a key component of our MCCL framework. For a given image \( I \), this loss minimizes the cosine similarity between foreground synergy features associated with different semantic categories, promoting inter-category feature disentanglement. It is formally defined as:
\begin{equation}
L_{\mathrm{sep}}(\mathcal{F}_I) = - \frac{1}{N_I^{\mathrm{pair}}}
\sum_{\substack{c_j, c_k \in \mathcal{C}_I\\ j \neq k}}
\log \left( 1 - \mathrm{sim}(f_{c_j}, f_{c_k}) \right),
\end{equation}
where \( f_{c_j}, f_{c_k} \in \mathcal{F}_I \) are the foreground synergy features corresponding to semantic categories \( c_j \) and \( c_k \) in image \( I \), and \( N_I^{\mathrm{pair}} = N_I(N_I - 1) \) denotes the number of ordered category pairs with \( j \neq k \).

Based on the description above, we define the total loss \( L_{\mathrm{total}} \) as a weighted sum of all components, with corresponding hyperparameters \( \lambda_1 \), \( \lambda_2 \), \( \lambda_3 \), and \( \lambda_4 \): 
\begin{equation}
L_{\mathrm{total}} = \lambda_1 L_{\mathrm{align}} + \lambda_2 L_{\mathrm{cont}} + \lambda_3 L_{\mathrm{back}} + \lambda_4 L_{\mathrm{sep}}.
\end{equation}

%% file: sec/4_experiment.tex

\section{Experiments}


\subsection{Experiment Setup}

\textbf{Training datasets.} We use the public conceptual-12m (CC12M)~\cite{cc12m} as training dataset. Utilizing the NLP functions from the SpaCy library, nouns and noun phrases are extracted from these captions. We resize the figures to the same pixel size of \(224 \times 224\) for training.  Data pre-processing details are provided in the appendix.

\textbf{Evaluating datasets.} To evaluate the open vocabulary semantic segmentation performance of our method, we test it on six commonly used challenging datasets: PASCAL VOC (VOC)~\cite{voc}, Pascal Context (Context)~\cite{context}, City Scapes (City)~\cite{cityspace} , ADE20K (ADE)~\cite{ade}, COCO Object (Object) and COCO Stuff (Stuff)~\cite{coco,stuff}. Also, it should be noted that evaluations on VOC and Object treat unlabeled regions as an explicit category \textit{background}, whereas those on Context, Stuff, City and ADE focus solely on labeled categories.

\textbf{Baselines.} We compare our proposed method, SynSeg, an end-to-end lightweight solution against eight classic and latest weakly-supervised open-vocabulary semantic segmentation approaches: GroupViT~\cite{groupvit}, ViewCo~\cite{ren2023viewco}, CoCu~\cite{cocu}, OVSegmentor~\cite{ov-seg}, TCL~\cite{tcl}, CoDe~\cite{code}, MGCA~\cite{MGCA} and S-Seg\cite{s-seg}. All baseline methods (including ours) are assumed to use a pre-trained ViT-B/16 vision backbone if available. These methods may use extra training datasets such as CC3M~\cite{cc3m}, YFCC14M~\cite{YFCC} and RedCaps12M~\cite{redcaps}.

\textbf{Implementation Details.\label{training details}} We use CLIP Vit-B/16 model as the encoder. The decoder follows the architecture of CLIPSeg~\cite{clipseg}. Training details are provided in the appendix. Following prior work DPSeg~\cite{dpseg}, we also adopt a multi-scale inference strategy to better handle high-resolution images during evaluation. Detailed settings are provided in the appendix.

\subsection{Main Results}
Tab.~\ref{tab:ovss-results} reports the mean Intersection-over-Union (mIoU) scores for each method on six representative benchmarks. SynSeg achieves the best overall performance, with an average mIoU of \textbf{36.5\%}, outperforming all baselines by a significant margin. To better understand, SynSeg achieves higher accuracy than SOTA baselines across all reported benchmarks, with gains ranging from 0.6\% up to 8.9\% mIoU scores.

\subsection{Visual Effects}
We present an example of semantic activation maps and the corresponding segmentation masks predicted by our model under multiple threshold settings, as illustrated in Fig.~\ref{fig:sample}. The segmentation results remain visually consistent across a wide range of thresholds, nearly from 0.1 to 0.6, particularly for prominent object categories such as person and taxi. This indicates that the model has an accurate semantic localization ability so that it produces clear boundaries. 

\begin{table}[htbp]
\vspace{-1em}
\centering
\caption{Ablation study on the training objectives.}
\setlength{\tabcolsep}{4pt}
\begin{tabular}{cccc|ccc}
\toprule
\multicolumn{4}{c|}{Loss} & \multicolumn{3}{c}{MIoU(\%)} \\
 $L_{cont}$ &$L_{align}$ & $L_{back}$ & $L_{sep}$  &\textbf{VOC} &\textbf{Context} &\textbf{City} \\
\midrule
\checkmark  & \checkmark & \checkmark & \checkmark &\textbf{60.5} &\textbf{43.8}  &\textbf{35.8} \\
\midrule
$\checkmark$ &$\times$  & $\times$  & $\times$  & 33.0 & 25.8  &  28.4  \\  

$\checkmark$ & $\checkmark$ &  $\times$  & $\times$   &50.7 & 28.4  &26.0\\ 

$\checkmark$  & $\checkmark$  &  $\checkmark$ &$\times$ & 54.6 &35.7 &30.1\\ 

$\times$ &  $\checkmark$ &$\checkmark$ &$\checkmark$   & 59.8 & 43.5 & 35.7 \\ 

\bottomrule
\end{tabular}

\label{tab:ablation}
\vspace{-2em}
\end{table}

\subsection{Ablation Study\label{ablation}}
We examine the effectiveness of the four training objectives by evaluating their impact on the segmentation performance across three datasets: Context, Stuff and City. The results, presented in Tab.~\ref{tab:ablation}, show that the combination of all four losses yields the best performance.


%% file: sec/5_conclusion.tex
\section{Conclusion}
We propose a novel approach, SynSeg, for end-to-end open-vocabulary semantic segmentation (OVSS), integrating Feature Synergy Structure (FSS) with Multi-Category Contrastive Learning (MCCL). Experiments on six benchmarks demonstrate the effectiveness of SynSeg, which consistently outperforms existing weakly-supervised baselines in OVSS, achieving state-of-the-art performance. 

%% file: sec/6_appendix.tex
\appendix

In appendix, we provide data pre-processing details, further experimental results, and other supplementary materials.

\subsection{Data Pre-Processing}
\label{appendix:preprocessing}

To construct high-quality noun-image pairs from the large-scale CC12M dataset for contrastive learning, we design a two-stage pipeline to extract and filter noun phrases from image captions.

\subsubsection{Stage 1: Noun Phrase Extraction}

We first extract noun phrases from each caption using the \texttt{spaCy} NLP library. Given a caption $x$, we extract syntactic noun chunks $\mathcal{N}(x) = \{n_1, n_2, \dots, n_k\}$ and retain those that contain at least one noun token (with \texttt{pos\_} tag \texttt{NOUN}) to ensure semantic relevance. This step eliminates phrases dominated by modifiers or non-content words.

The result is a mapping from each image to a set of candidate noun phrases:
\[
\texttt{Image}_{i} \rightarrow \texttt{['dog', 'park', 'south', ...]}
\]

The extraction process is summarized in Algorithm~\ref{alg:extract}.

\begin{algorithm}[b]
\caption{Noun Phrase Extraction from Captions}
\label{alg:extract}
\textbf{Input}: Caption file with image-caption pairs \\
\textbf{Output}: Mapping from image to noun phrases
\begin{algorithmic}[1]
\FOR{each line in the caption file}
    \STATE Parse caption using \texttt{spaCy}
    \STATE noun\_phrases $\leftarrow$ []
    \FOR{each chunk in parsed.noun\_chunks}
        \STATE Filter tokens with \texttt{pos\_} == \texttt{NOUN}
        \IF{at least one noun is retained}
            \STATE Join tokens and append to noun\_phrases
        \ENDIF
    \ENDFOR
\ENDFOR
\STATE \textbf{return} image $\rightarrow$ noun\_phrases
\end{algorithmic}
\end{algorithm}

\subsubsection{Stage 2: Generic Phrase Filtering}

To enhance the grounding quality of the extracted noun phrases, we further filter out overly generic or non-referential terms such as ``reflection'', ``southwest'' or ``background'' that lack clear visual correspondence. This is achieved by performing partial string matching against a manually defined exclusion list $\mathcal{E}$. A phrase $p$ is retained only if it contains no substring from $\mathcal{E}$:
\[
\forall e \in \mathcal{E},\quad e \notin p
\]

The full filtering procedure is outlined in Algorithm~\ref{alg:filter}.

\begin{algorithm}[b]
\caption{Generic Phrase Filtering}
\label{alg:filter}
\textbf{Input}: List of noun phrases, exclusion list $\mathcal{E}$ \\
\textbf{Output}: Filtered noun phrase list
\begin{algorithmic}[1]
\FOR{each noun phrase $p$ in the list}
    \IF{$\forall e \in \mathcal{E}, \quad e \notin p$}
        \STATE Keep $p$
    \ELSE
        \STATE Discard $p$
    \ENDIF
\ENDFOR
\STATE \textbf{return} filtered noun phrases
\end{algorithmic}
\end{algorithm}

The final output is a cleaned set of noun-image pairs that serve as anchors for downstream contrastive training.

\subsection{Multi-Scale Inference Strategy}

CLIP-based vision encoders only support fixed-resolution inputs (e.g., 224×224), which may lead to substantial information loss when processing high-resolution images. To mitigate this issue, we adopt a modified multi-scale inference strategy inspired by DPSeg~\cite{dpseg}. Specifically, each input image is evenly divided into four non-overlapping patches, which are individually resized to 224×224. In addition, the original image is resized to the same resolution. These five views are grouped into a single batch and jointly processed during inference. This strategy effectively preserves fine-grained spatial details while remaining compatible with fixed-resolution CLIP encoders.

\subsection{Additional Visual Effects}

\begin{figure*}[htbp]
\centerline{\includegraphics[scale=0.9]{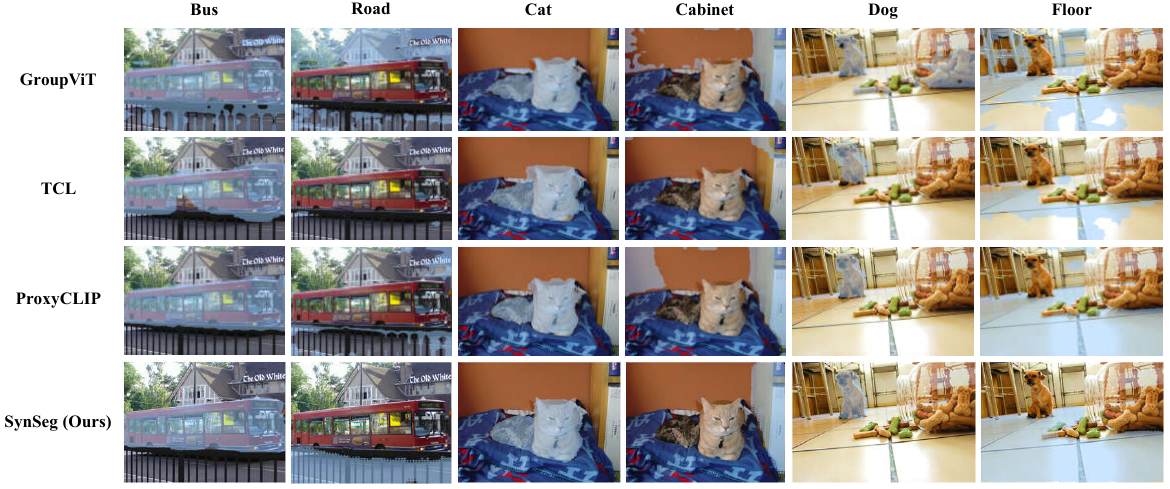}}
\caption{\textbf{Segmentation visual comparisons}. The light blue regions indicate the segmentation predictions. The baselines' results are visually compared with our method, SynSeg. The examples are selected from the Context dataset~\cite{context}.}
\label{fig:sample2}
\end{figure*}

In addition to the results in the main manuscript, we conduct visual comparisons between our method and existing methods. Our comparison includes weakly-supervised approaches GroupViT~\cite{groupvit} and TCL~\cite{tcl}, and a strong training-free baseline ProxyCLIP~\cite{proxy}. ProxyCLIP utilizes a large-scale pretrained Vision Foundation Model (VFM) DINO~\cite{dino} for OVSS task. As shown in Fig.~\ref{fig:sample2}, our method consistently demonstrates more accurate semantic localization and finer-grained segmentation. The predicted segmentations from our model are better aligned with the object boundaries and exhibit fewer false positives.

In general, our segmentation results show superior performance in both salient and background categories. For salient objects such as bus, cat, and dog, our model produces tighter and more precise boundaries, with reduced over-segmentation compared to others which only ProxyCLIP performs competitively. For background regions such as floor and road, our method significantly outperforms all baselines. ProxyCLIP tends to misclassify visually similar areas as background categories, while GroupViT and TCL often fail to produce coherent masks in these regions. These observations further demonstrate the robustness of our method.

\subsection{Experimental Setup}
\label{appendix:setup}

\subsubsection{FiLM}
Feature-wise Linear Modulation (FiLM) is a cross-modal fusion module. Specifically, the FiLM module contains a small learnable MLP that takes the text features as input and outputs channel-wise scaling and shifting parameters. These parameters are then directly applied to the visual features to generate the conditional visual vectors.

\subsubsection{Experimental Enviroment}
In order to satisfy the requirments of the reproducibility checklist, we report our experimental setup in detail here. All experiments in this paper were conducted on a dedicated server equipped with 8$\times$NVIDIA RTX 4090 GPUs (24GB each) and a single Intel Xeon Platinum 8358P CPU. To ensure reproducibility, we fix the random seed to 42 in all experiments. And each experiment (including training and evaluation) was run independently on a single RTX 4090 GPU, as our model and training recipes are sufficiently lightweight. All reported results are averaged over five independent runs. 

\subsubsection{Hyperparameter Configuration}
The hyperparameter settings for the main experiments are as follows: the loss weights are set to $\lambda_1 = 1 \times 10^{-1}$, $\lambda_2 = 1 \times 10^{-3}$, $\lambda_3 = 1 \times 10^{-2}$, and $\lambda_4 = 1$; the learning rate is $1 \times 10^{-5}$ with a weight decay of $1 \times 10^{-2}$; the batch size is 256; the model is trained for 10,000 iterations using the AdamW optimizer. These values are used consistently unless otherwise specified. Although CC12M contains a substantially larger number of image–text pairs, we did not utilize the full dataset for training. Empirically, the model converged to a saturated performance after approximately 10,000 training iterations, and further training did not yield noticeable improvements.

In implementation, \(L_{\mathrm{cont}}\) is activated only for foreground-background pairs whose cosine similarity exceeds a preset threshold 0.1. The pairs that are already sufficiently separated are excluded to avoid unnecessary repulsion and stabilize optimization.

\subsection{Dataset Details}
\label{appendix:datasets}

We evaluate our method on six widely used open-vocabulary semantic segmentation benchmarks: PASCAL VOC~\cite{voc}, Pascal Context~\cite{context}, Cityscapes~\cite{cityspace}, COCO Object, COCO Stuff~\cite{coco,stuff}, and ADE20K (ADE)~\cite{ade}. These datasets collectively cover a diverse range of semantic granularity and scene complexity, which makes them suitable for comprehensive evaluation under open vocabuary conditions. Table~\ref{tab:dataset-summary} summarizes their statistics.

\begin{table}[h]
\centering
\caption{Summary of dataset statistics and evaluation settings.}
\small
\begin{tabular}{lccc}
\toprule
Dataset & \#Categories & Labels/Image & Background\\
\midrule
Pascal VOC     & 21   & 2.5 & Yes \\
Pascal Context & 59   & 4.8 & No  \\
Cityscapes     & 19   & 12.0 & No  \\
COCO Object    & 81   & 7.2 & Yes \\
COCO Stuff     & 171  & 8.4 & No  \\
ADE            & 151  & 9.5 & No \\
\bottomrule
\end{tabular}

\label{tab:dataset-summary}
\end{table}

\subsection{LLM Usage}

We declare that we only used the Large Language Models (LLMs) for language polishing and grammar correction during the writing of this manuscript. All research content, ideas, experiments and conclusions are solely produced by the authors. In summary, we take full responsibility for the entire content of this paper.

%% file: main.bbl
\begin{thebibliography}{10}

\bibitem{groupvit}
Jiarui Xu, Shalini De~Mello, Sifei Liu, Wonmin Byeon, Thomas Breuel, Jan Kautz, and Xiaolong Wang,
\newblock ``Groupvit: Semantic segmentation emerges from text supervision,''
\newblock in {\em CVPR}, June 2022, pp. 18134--18144.

\bibitem{pacl}
Jishnu Mukhoti, Tsung-Yu Lin, Omid Poursaeed, Rui Wang, Ashish Shah, Philip~H.S. Torr, and Ser-Nam Lim,
\newblock ``Open vocabulary semantic segmentation with patch aligned contrastive learning,''
\newblock in {\em CVPR}, June 2023, pp. 19413--19423.

\bibitem{tcl}
Junbum Cha, Jonghwan Mun, and Byungseok Roh,
\newblock ``Learning to generate text-grounded mask for open-world semantic segmentation from only image-text pairs,''
\newblock in {\em CVPR}, 2023.

\bibitem{proxy}
Mengcheng Lan, Chaofeng Chen, Yiping Ke, Xinjiang Wang, Litong Feng, and Wayne Zhang,
\newblock ``Proxyclip: Proxy attention improves clip for open-vocabulary segmentation,''
\newblock in {\em ECCV}, Berlin, Heidelberg, 2024, p. 70–88, Springer-Verlag.

\bibitem{cocu}
Yun Xing, Jian Kang, Aoran Xiao, Jiahao Nie, Shao Ling, and Shijian Lu,
\newblock ``Rewrite caption semantics: Bridging semantic gaps for language-supervised semantic segmentation,''
\newblock in {\em NeurIPS}, 2023.

\bibitem{code}
Ji-Jia Wu, Andy Chia-Hao Chang, Chieh-Yu Chuang, Chun-Pei Chen, Yu-Lun Liu, Min-Hung Chen, Hou-Ning Hu, Yung-Yu Chuang, and Yen-Yu Lin,
\newblock ``Image-text co-decomposition for text-supervised semantic segmentation,''
\newblock in {\em CVPR}, June 2024, pp. 26794--26803.

\bibitem{ov-seg}
Jilan Xu, Junlin Hou, Yuejie Zhang, Rui Feng, Yi~Wang, Yu~Qiao, and Weidi Xie,
\newblock ``Learning open-vocabulary semantic segmentation models from natural language supervision,''
\newblock in {\em CVPR}, 2023, pp. 2935--2944.

\bibitem{cat-seg}
Seokju Cho, Heeseong Shin, Sunghwan Hong, Anurag Arnab, Paul~Hongsuck Seo, and Seungryong Kim,
\newblock ``Cat-seg: Cost aggregation for open-vocabulary semantic segmentation,''
\newblock in {\em CVPR}, June 2024, pp. 4113--4123.

\bibitem{openseg}
Golnaz Ghiasi, Xiuye Gu, Yin Cui, and Tsung-Yi Lin,
\newblock ``Scaling open-vocabulary image segmentation with image-level labels,''
\newblock in {\em ECCV}, Berlin, Heidelberg, 2022, p. 540–557, Springer-Verlag.

\bibitem{s-seg}
Zihang Lai,
\newblock ``Exploring simple open-vocabulary semantic segmentation,''
\newblock in {\em CVPR}, 2025, pp. 30221--30230.

\bibitem{ren2023viewco}
Pengzhen Ren, Changlin Li, Hang Xu, Yi~Zhu, Guangrun Wang, Jianzhuang Liu, Xiaojun Chang, and Xiaodan Liang,
\newblock ``Viewco: Discovering text-supervised segmentation masks via multi-view semantic consistency,''
\newblock in {\em ICLR}, 2023.

\bibitem{stuff}
Holger Caesar, Jasper Uijlings, and Vittorio Ferrari,
\newblock ``Coco-stuff: Thing and stuff classes in context,''
\newblock in {\em CVPR}, 2018, pp. 1209--1218.

\bibitem{coco}
Tsung-Yi Lin, Michael Maire, Serge Belongie, James Hays, Pietro Perona, Deva Ramanan, Piotr Doll{\'a}r, and C.~Lawrence Zitnick,
\newblock ``Microsoft coco: Common objects in context,''
\newblock in {\em ECCV}, 2014, pp. 740--755.

\bibitem{clip}
Alec Radford, Jong~Wook Kim, Chris Hallacy, Aditya Ramesh, Gabriel Goh, Sandhini Agarwal, Girish Sastry, Amanda Askell, Pamela Mishkin, Jack Clark, Gretchen Krueger, and Ilya Sutskever,
\newblock ``Learning transferable visual models from natural language supervision,''
\newblock in {\em ICML}. 18--24 Jul 2021, vol. 139, pp. 8748--8763, PMLR.

\bibitem{denseclip}
Yongming Rao, Wenliang Zhao, Guangyi Chen, Yansong Tang, Zheng Zhu, Guan Huang, Jie Zhou, and Jiwen Lu,
\newblock ``Denseclip: Language-guided dense prediction with context-aware prompting,''
\newblock in {\em CVPR}, 2022.

\bibitem{mm_survey}
Tadas Baltrušaitis, Chaitanya Ahuja, and Louis-Philippe Morency,
\newblock ``Multimodal machine learning: A survey and taxonomy,''
\newblock {\em IEEE Transactions on Pattern Analysis and Machine Intelligence}, vol. 41, no. 2, pp. 423--443, 2019.

\bibitem{WENJUN}
Wenjun Lyu, Xiaolong Jin, Haotian Wang, Yiwei Song, Shuai Wang, Yunhuai Liu, Tian He, and Desheng Zhang,
\newblock ``Towards workload-constrained efficient order assignment in last-mile delivery,''
\newblock {\em IEEE Transactions on Mobile Computing}, vol. 24, no. 2, pp. 557--570, 2025.

\bibitem{align}
Junnan Li, Ramprasaath Selvaraju, Akhilesh Gotmare, Shafiq Joty, Caiming Xiong, and Steven Chu~Hong Hoi,
\newblock ``Align before fuse: Vision and language representation learning with momentum distillation,''
\newblock in {\em NeurIPS}, 2021, vol.~34, pp. 9694--9705.

\bibitem{blip}
Junnan Li, Dongxu Li, Caiming Xiong, and Steven Hoi,
\newblock ``{BLIP}: Bootstrapping language-image pre-training for unified vision-language understanding and generation,''
\newblock in {\em ICML}. 17--23 Jul 2022, vol. 162 of {\em Proceedings of Machine Learning Research}, pp. 12888--12900, PMLR.

\bibitem{vinvl}
Pengchuan Zhang, Xiujun Li, Xiaowei Hu, Jianwei Yang, Lei Zhang, Lijuan Wang, Yejin Choi, and Jianfeng Gao,
\newblock ``Vinvl: Revisiting visual representations in vision-language models,''
\newblock in {\em CVPR}, 2021, pp. 5575--5584.

\bibitem{film}
Ethan Perez, Florian Strub, Harm de~Vries, Vincent Dumoulin, and Aaron Courville,
\newblock ``Film: visual reasoning with a general conditioning layer,''
\newblock in {\em AAAI}. 2018, AAAI'18/IAAI'18/EAAI'18, AAAI Press.

\bibitem{maskclip}
Zheng Ding, Jieke Wang, and Zhuowen Tu,
\newblock ``Open-vocabulary universal image segmentation with maskclip,''
\newblock in {\em ICML}. 2023, JMLR.org.

\bibitem{trident}
Yuheng Shi, Minjing Dong, and Chang Xu,
\newblock ``Harnessing vision foundation models for high-performance, training-free open vocabulary segmentation,''
\newblock {\em arXiv preprint arXiv:2411.09219}, 2024.

\bibitem{kim2024d}
Chanyoung Kim, Dayun Ju, Woojung Han, Ming-Hsuan Yang, and Seong~Jae Hwang,
\newblock ``Distilling spectral graph for object-context aware pen-vocabulary semantic segmentation,''
\newblock in {\em CVPR}, 2025.

\bibitem{dpseg}
Ziyu Zhao, Xiaoguang Li, Linjia Shi, Nasrin Imanpour, and Song Wang,
\newblock ``Dpseg: Dual-prompt cost volume learning for open-vocabulary semantic segmentation,''
\newblock in {\em CVPR}, 2025, pp. 25346--25356.

\bibitem{dino}
Cijo Jose, Théo Moutakanni, Dahyun Kang, Federico Baldassarre, Timothée Darcet, Hu~Xu, Daniel Li, Marc Szafraniec, Michaël Ramamonjisoa, Maxime Oquab, Oriane Siméoni, Huy~V. Vo, Patrick Labatut, and Piotr Bojanowski,
\newblock ``Dinov2 meets text: A unified framework for image- and pixel-level vision-language alignment,'' 2024.

\bibitem{sdiffusion}
Robin Rombach, Andreas Blattmann, Dominik Lorenz, Patrick Esser, and Bj\"orn Ommer,
\newblock ``High-resolution image synthesis with latent diffusion models,''
\newblock in {\em CVPR}, June 2022, pp. 10684--10695.

\bibitem{ccam}
Jinheng Xie, Jianfeng Xiang, Junliang Chen, Xianxu Hou, Xiaodong Zhao, and Linlin Shen,
\newblock ``C2am: Contrastive learning of class-agnostic activation map for weakly supervised object localization and semantic segmentation,''
\newblock in {\em CVPR}, June 2022, pp. 989--998.

\bibitem{cc12m}
Soravit Changpinyo, Piyush Sharma, Nan Ding, and Radu Soricut,
\newblock ``Conceptual 12m: Pushing web-scale image-text pre-training to recognize long-tail visual concepts,'' 2021.

\bibitem{voc}
Mark Everingham, Luc Gool, Christopher~K. Williams, John Winn, and Andrew Zisserman,
\newblock ``The pascal visual object classes (voc) challenge,''
\newblock {\em Int. J. Comput. Vision}, vol. 88, no. 2, pp. 303–338, June 2010.

\bibitem{context}
Roozbeh Mottaghi, Xianjie Chen, Xiaobai Liu, Nam-Gyu Cho, Seong-Whan Lee, Sanja Fidler, Raquel Urtasun, and Alan Yuille,
\newblock ``The role of context for object detection and semantic segmentation in the wild,''
\newblock in {\em CVPR}, 2014, pp. 891--898.

\bibitem{cityspace}
Marius Cordts, Mohamed Omran, Sebastian Ramos, Timo Rehfeld, Markus Enzweiler, Rodrigo Benenson, Uwe Franke, Stefan Roth, and Bernt Schiele,
\newblock ``The cityscapes dataset for semantic urban scene understanding,''
\newblock in {\em CVPR}, 2016.

\bibitem{ade}
Bolei Zhou, Hang Zhao, Xavier Puig, Tete Xiao, Sanja Fidler, Adela Barriuso, and Antonio Torralba,
\newblock ``Semantic understanding of scenes through the ade20k dataset,''
\newblock {\em Int. J. Comput. Vision}, vol. 127, no. 3, pp. 302–321, Mar. 2019.

\bibitem{MGCA}
Yajie Liu, Pu~Ge, Guodong Wang, Qingjie Liu, and Di~Huang,
\newblock ``Multi-grained contrastive learning for text-supervised open-vocabulary semantic segmentation,''
\newblock {\em ACM Trans. Multimedia Comput. Commun. Appl.}, vol. 21, no. 3, Feb. 2025.

\bibitem{cc3m}
Piyush Sharma, Nan Ding, Sebastian Goodman, and Radu Soricut,
\newblock ``Conceptual captions: A cleaned, hypernymed, image alt-text dataset for automatic image captioning,''
\newblock in {\em Proceedings of the 56th Annual Meeting of the Association for Computational Linguistics}, Melbourne, Australia, July 2018, pp. 2556--2565, Association for Computational Linguistics.

\bibitem{YFCC}
Bart Thomee, David~A. Shamma, Gerald Friedland, Benjamin Elizalde, Karl Ni, Douglas Poland, Damian Borth, and Li-Jia Li,
\newblock ``Yfcc100m: the new data in multimedia research,''
\newblock {\em Commun. ACM}, vol. 59, no. 2, pp. 64–73, Jan. 2016.

\bibitem{redcaps}
Karan Desai, Gaurav Kaul, Zubin Aysola, and Justin Johnson,
\newblock ``{RedCaps: Web-curated image-text data created by the people, for the people},''
\newblock in {\em NeurIPS Datasets and Benchmarks}, 2021.

\bibitem{clipseg}
Timo L\"uddecke and Alexander Ecker,
\newblock ``Image segmentation using text and image prompts,''
\newblock in {\em CVPR}, June 2022, pp. 7086--7096.

\end{thebibliography}
